\documentclass[runningheads]{llncs}
\usepackage{marvosym}
\usepackage{graphicx}
\usepackage{nicefrac}       
\usepackage{xcolor}

\usepackage{microtype}      
\usepackage{lipsum}         
\usepackage{booktabs}
\usepackage{doi}
\usepackage{listings}
\usepackage{verbatim}
\usepackage{amssymb}
\usepackage{amsfonts}
\usepackage{algorithmic}
\usepackage{algorithm2e}
\usepackage{mathrsfs}
\usepackage{float}
\usepackage{textcomp}
\usepackage{animate}
\usepackage{tikz}
\usepackage{multirow}
\usepackage{amsmath}
\usepackage{times}
\usepackage{helvet}
\usepackage{courier}
\usepackage{comment}
\usepackage{latexsym}
\usepackage{mathtools}
\usepackage{subcaption}

\usetikzlibrary{positioning, arrows}

\begin{document}
\title{Explaining Reinforcement Learning: A Counterfactual Shapley Values Approach }
%
%
\author{Yiwei Shi$^1$, Qi Zhang$^2$, Kevin McAreavey$^1$, Weiru Liu$^1$
}
\authorrunning{Yiwei et al.}
\institute{$^1$ School of Engineering Mathematics and Technology, University of Bristol\\ $^2$ School of Electronic and Information Engineering, Tongji University\\
\email{~}}
\maketitle              
\begin{abstract}
This paper introduces a novel approach Counterfactual Shapley Values (CSV), which enhances explainability in reinforcement learning (RL) by integrating counterfactual analysis with Shapley Values. The approach aims to quantify and compare the contributions of different state dimensions to various action choices. To more accurately analyze these impacts, we introduce new characteristic value functions, the ``Counterfactual Difference Characteristic Value" and the ``Average Counterfactual Difference Characteristic Value." These functions help calculate the Shapley values to evaluate the differences in contributions between optimal and non-optimal actions. Experiments across several RL domains, such as GridWorld, FrozenLake, and Taxi, demonstrate the effectiveness of the CSV method. The results show that this method not only improves transparency in complex RL systems but also quantifies the differences across various decisions.

\end{abstract}
\section{Introduction}
 Reinforcement learning (RL) has applications in autonomous navigation \cite{chen2021interpretable}, healthcare \cite{liu2019deep}, financial strategy optimization \cite{jiang2017deep}, and smart city management \cite{kumar2021deep}. However, the key to achieving widespread adoption of these applications lies in overcoming the challenge of explainability in RL models. the inability to adequately explain RL models could provoke ethical and regulatory issues and meet with resistance from potential users, collectively impeding the extensive application and further advancement of RL technology.

Research in explainable reinforcement learning is primarily focused on two approaches: \textit{intrinsic interpretability} and \textit{post-hoc interpretability}. Intrinsic interpretability strategies enhance model transparency by simplifying the model's structure, often at the expense of reducing the model's performance. For instance, decision tree models, which represent intrinsic interpretability, visually demonstrate decision pathways through their tree-like structure but may fail to capture more complex data patterns. In contrast, post-hoc interpretability methods treat the model as a black box and reveal the decision-making logic by analyzing the relationships between inputs and outputs, thus preserving the model's complexity. \textit{Counterfactuals} and \textit{Shapley Values} are both post-hoc explanation methods that provide insights without intervening in the internal structure of the model.

Shapley Value is a method for measuring the contribution of individuals to the total payoff in cooperative games. An application of Shapley values to Explainable Artificial Intelligence (XAI), each feature is regarded as a ``player", while the model's predictive outcome is considered the ``total payoff" \cite{lundberg2017unified}. The Shapley Value assesses the average contribution of each feature to the predictive outcome by considering all possible combinations of features. 

Investigations employing Shapley Values are remarkably limited across both multi-agent reinforcement learning (MARL) and single-agent in RL. However, MARL research demonstrates a relatively richer engagement with Shapley value applications, which are principally concentrated across three aspects. Firstly, \textit{Value Decomposition},
\cite{NEURIPS2022_27985d21}  integrates Shapley Value theory with multi-agent Q-learning (SHAQ) to address value decomposition in global reward games, where Shapley Values play a critical role in ensuring equitable distribution of rewards based on individual agent contributions.
Secondly, \textit{Credit Assignment},
\cite{wang2020shapley} introduces the Shapley Q-value within MARL (SQDDPG) to reallocate global rewards among agents, leveraging Shapley Values to quantify and fairly distribute the rewards reflecting each agent's marginal contribution to the collective success.
Lastly, \textit{Model Explanation}, \cite{Heuillet2022CollectiveXA} the use of Shapley Values to explain cooperative strategies and individual contributions in multi-agent reinforcement learning, approximating Shapley Values via Monte Carlo sampling to reduce computational costs. Shapley Values offer a way to interpret complex multi-agent interactions, making it clearer how each agent's decisions and actions contribute to the overall dynamics and results of the system.

To our knowledge, only the SVERL in \cite{beechey2023explaining}, within the scope of RL research excluding MARL, utilizes Shapley Values to explain the decision-making process in reinforcement learning. Although SVERL provides a method for understanding and explaining the contribution of state features in the decision-making process of reinforcement learning agents, it has limitations in explaining specific action choices. The core of SVERL lies in analyzing the contribution of state features to the expected returns of an agent, rather than directly investigating why action $a$ is chosen over action $b$ in the same state, and it does not offer a mechanism to quantify the difference between these action choices. This means that while SVERL can help us indirectly understand how certain features influence an agent's decisions, it cannot directly answer why an agent prefers a specific action, nor can it compare the merits of different actions. This limitation is not unique to SVERL but is a challenge faced by all methods that rely on Shapley Values to explain the decision-making process in reinforcement learning.

Counterfactuals can effectively address some of the limitations associated with explaining specific action choices in RL. By comparing the utility (Long-Term Expected Return) of the actual action with those of the counterfactual (alternative) action, counterfactuals can help clarify why an agent prefers one action over another and quantify the impact of different actions.

In reinforcement learning, two main generative explanation methods by counterfactuals - one based on \textbf{deep generative models} and the other on \textbf{generating counterfactual states} - share a common limitation: the difficulty in quantifying differences between counterfactual instances. While methods based on deep generative models \cite{Olson2021Counterfactual} can create realistic ``what if" scenarios, comparing these instances in a high-dimensional latent space is challenging, as conventional distance metrics may not apply. On the other hand, methods that focus on minor state adjustments to guide different decisions also face challenges in quantifying the actual impact of these subtle changes on agent behavior, since even small modifications can have widespread effects in complex RL environments. Therefore, although these methods provide valuable insights into the decision-making process in RL, their difficulty in quantifying differences between counterfactual instances limits their application scope and the depth of their explanations.

To address the challenges of quantifying differences in reinforcement learning explanations, we develop a novel approach called ``Counterfactual Shapley Value". While previous research has explored the integration of counterfactual explanations with Shapley Values, such as addressing credit assignment problems in MARL in \cite{li2021shapley} or providing explanations in supervised learning \cite{albini2022counterfactual}, the former develops a novel method to assign credits in multi-agent systems using counterfactual thinking and Shapley values, which quantifies each agent's contribution by considering what would happen if certain agents were absent, allowing for a fair and accurate assessment of each agent’s impact on the collective outcome, the latter uses counterfactual scenarios to enhance Shapley value explanations, providing insights on how changes to input features could influence model predictions. It makes explanations more actionable by showing which features to adjust to achieve desired outcomes. Our work is fundamentally different. For the first time, we have applied the combination of counterfactual reasoning and Shapley Values to the domain of reinforcement learning, introducing the innovative CSV approach. This method not only retains the benefits of traditional Shapley Values in quantitative analysis but also enhances the understanding of decision-making by examining specific hypothetical scenarios, such as ``What would happen if a specific action were not taken?". Our approach brings a more transparent and interpretable perspective to decision-making in reinforcement learning, significantly alleviating the limitations of previous explanatory methods.

In this paper, our research contributions are threefold: Firstly, we have introduced an innovative method called Counterfactual Shapley Value, which can precisely reveal the changes and differences in the contributions of state features to different actions within reinforcement learning. Secondly, we have designed two novel characteristic value functions - \textit{the Counterfactual Difference Characteristic Value} and \textit{the Average Counterfactual Difference Characteristic Value}, which incorporate counterfactual mechanisms for calculating Shapley Values, offering us new avenues for computation. Lastly, we have tested and analyzed this method across multiple RL environments, where the quantified results from CSV have allowed us to explain the behavior of agents, thereby demonstrating the effectiveness of our approach.

\section{PRELIMINARIES}
\subsection{Markov Decision Process (MDP)}
In the domain of reinforcement learning, an agent's interaction with the environment is conceptualized as a Markov Decision Process (MDP), denoted by \( MDP = (S, A, P, R, \gamma\)). Here, \( S \) is the set of all possible states, \( A \) is the set of actions, \( P \) is the state transition probabilities \( P(s'|s,a) : S \times A \times S \rightarrow [0,1]\), \( R(s,a) : S \times A \rightarrow \mathbb{R}\) represents the reward function \( R(s, a) \), \( \gamma \in [0,1]\) denotes the discount factor the cumulative reward \( G_t \) represents the sum of discounted future rewards starting from time \( t \). It is mathematically defined as \( G_t = \sum_{k=0}^{\infty} \gamma^k R_{t+k+1} \), with \( R_{t+k+1} \) signifying the immediate reward received at time \( t+k+1 \). The role of cumulative reward is fundamental, serving as a critical metric to assess and guide the agent's performance. The agent's objective is to discover a policy \( \pi : S \times A \rightarrow [0, 1] \) that assigns probabilities to actions in each state, with the aim of optimizing the cumulative expected reward over time. The state-value function \( V^\pi(s) = \mathbb{E}^\pi \left[ G_t \mid S_t = s \right] \) computes the expected return from starting in state \( s \) under policy \( \pi \), while the action-value function \( Q^\pi(s, a) = \mathbb{E}^\pi \left[ G_t \mid S_t = s, A_t = a \right] \) assesses the expected return after taking action \( a \) in state \( s \) under the same policy. The optimal policy, therefore, is defined by maximizing these functions, \( V^*(s) \) and \( Q^*(s, a) \), to achieve the highest possible expected returns from all states and actions.

\subsection{Shapley Values in RL}
In the domain of cooperative game theory, Shapley Values, originally proposed by \cite{Shapley1953}, serve as a pivotal solution concept for the equitable allocation of collective gains or losses among participating entities, predicated on their individual contributions to the cumulative outcome. When transposed to the Reinforcement Learning (RL) framework, this concept facilitates an analytical assessment of the individual contributions of state features to the decision-making efficacy of an RL agent.

In RL, let \( F = \{0, 1, \ldots, n-1\} \) represent the set of indices corresponding to the comprehensive enumeration of state features within the state space \( S \). The state space itself is structured into \( n \) discrete dimensions, expressed as \( S = S^{(0)} \times S^{(1)} \times \ldots \times S^{(n-1)} \), where each \( S^{(i)} \) captures a distinct aspect of the environmental dynamics. Within this framework, each state is an ordered set, defined by \( s = \{ s^{(i)} | s^{(i)} \in S^{(i)} , i \in F\} \). Let \( C \subseteq F \) denote a subset of state features, specifically excluding the feature indexed by \( i \). A partial observation of the state is then given by \( s^C = \{s^{(i)} | s^{(i)} \in S^{(i)}, i \in C\} \), which offers a perspective of the state based on a subset of observable features. The Shapley value \cite{Shapley1953} for a state feature \( i \), denoted as \( \phi_i \), quantifies the contribution of feature \( i \) and is mathematically expressed as follows:

\begin{equation} 
\phi_i = \sum_{C \subseteq F \setminus \{i\}} \frac{|C|!(|F| - |C| - 1)!}{|F|!} \cdot  \delta (i,C)
\end{equation}
where \(\delta (i,C) =  v(C \cup \{i\}) - v(C)\) quantifies the marginal contribution in the characteristic value function attributed to the incorporation of feature \(i\) into the subset \(C\), where \(v(C)\) represents the characteristic value  function assessed over the subset \(C\) excluding of feature \(i\), and \(v(C \cup \{i\})\) represents the characteristic value function with the inclusion of feature \(i\). This formulation enables the computation of the average marginal contribution of each state feature \(i\) across all conceivable coalitions of other features, thereby elucidating its significance in the decision-making processes of the RL agent.

\subsection{Characteristic Value Function}
In the field of machine learning, the integration of the characteristic value function (CVF) with Shapley Values provides a systematic method for quantifying the individual contributions of features towards predictive outcomes. This approach leverages principles from cooperative game theory, adapting them to clarify the significance of features within predictive models. Let \( f^F : S \rightarrow \mathbb{R} \) denote the characteristic function defined over the complete feature set \( F \) in state space $S$, which estimates the expected value of model predictions as \( v(F) := f^F(s) \). Correspondingly, for a subset \( {C} \subset {F} \), the characteristic function \( f^C : S^C \rightarrow \mathbb{R} \) assesses the expected prediction value \( v(C) := f^C(s^C) \), where \( S^C \) denotes the subspace restricted to the features in \( \mathcal{C} \). When all features are unknown, that is, no features are input into the model, it can be assumed that the feature set is an empty set (\( \emptyset \)). Then, the function's value is typically set to 0, denoted as \( v(\emptyset) = 0 \), indicating that the contribution of features is zero.

\section{COUNTERFACTUAL SHAPLEY VALUES}

In the XRL, it is imperative to gain a comprehensive understanding of the fundamental motivations underlying the decision-making processes of agents. While Shapley Values do not clarify the policy significance of action choices directly, the incorporation of counterfactual analysis enhances the ability to assess the potential effects of various actions in hypothetical scenarios, thereby offering deep insights into the policy selections of agents.

It is noteworthy that the value functions \( Q(s, a) \) and \( V(s) \), while pivotal in guiding agents towards optimal decision-making, also serve as \textit{utility functions} for computing Shapley Values, highlighting their significance in analyzing agent behavior. By employing these value functions as characteristic value function, a more precise quantitative analysis of the specific contributions of different dimensions in any given state can be achieved.

\subsection{Counterfactual Differences}

From the perspective of explaining  agent behavior, effectively comparing two distinct actions requires not only evaluating the immediate reward following action execution and the expected long-term return but also exploring counterfactual scenarios —``what would happen if a different action were chosen?" To facilitate this, we introduce the notion of counterfactual difference (CD), including both \textit{ action counterfactual differences} (\( \Delta Q \)) and \textit{state counterfactual differences} (\( \Delta V \)), to thoroughly analyze the impacts of various actions. The action counterfactual differences examines the difference in \( Q \) values between the actual action taken and a hypothetical alternative action in a given state, while the state counterfactual  differences focuses on the variance in \( V \) values for the new states resulting from these actions.

The action counterfactual differences, denoted as \( \Delta Q^{\pi}(s, a^*, a) \), is calculated by the comparison between the expected return of taking an optimal action \( a^* \) under the policy \( \pi \) in state \( s \), against the expected return of taking the other action \( a \) under the same policy \( \pi \) in the same state \( s \). This comparison yields the differences value, revealing the potential utility deviation resulting from an alternative action choice. \( \Delta Q^{\pi}(s, a^*, a) \) is calculated as follows:
\begin{equation}
\begin{split}
\Delta Q^{\pi}(s, a^*, a) = Q^{\pi}(s, a^*) - Q^{\pi}(s, a)
\end{split}
\end{equation}

\noindent where \( Q^{\pi}(s, a^*) \) represents the expected return of executing the optimal action \( a^* \) in state \( s \) under the policy \( \pi \), while \( Q^{\pi}(s, a) \) signifies the expected return of executing the counterfactual action \( a \) under the same policy \( \pi \) in the same state. Although the policy usually prioritizes executing what is assessed as the optimal action, the reward of non-optimal actions is also recorded during training to maintain a balance between exploration and exploitation. This forms the basis for a counterfactual analysis scenario, through which we can compare the expected return of non-optimal actions to optimal actions, thereby evaluating how different state features contribute to these two types of decisions.

The policy \( \pi \) can be either a ``fully learned policy'' or a ``partially learned policy''. In both types, the actions considered optimal are those with the highest Q values in each state. Typically, policies will choose these actions with the highest Q values for execution. A fully learned policy can always accurately execute these optimal choices. However, a partially learned policy, although theoretically aimed at maximizing rewards, may not always choose the best actions in practice. Since partially learned policies struggle to accurately calculate Q values to effectively predict long-term returns, it becomes difficult to assess how different state features contribute to decision-making. For these reasons, this study only considers policies that are fully learned and can execute accurately.

The state counterfactual differences, denoted as \( \Delta V^{\pi}(s^*, s') \), quantifies the variation in expected return associated with transitioning from the current state \( s \) to an optimal state \( s^* \) under the policy \( \pi \), in contrast to transitioning to a different state \( s' \) under the same policy \( \pi \). This differences highlights the expected return change resulting from adopting the optimal action compared to an alternative action from the same starting point. \( \Delta V^{\pi}(s^*, s') \) is calculated as follows:
\begin{equation}
\begin{split}
\Delta V^{\pi}(s^*, s') = V^{\pi}(s^*) - V^{\pi}(s')
\end{split}
\end{equation}

\textcolor{black}{
Although some states \( s' \) might only be reached through suboptimal actions \( a \) with very low probabilities after learned policy, these states still have definitive value functions, which reflect the expected rewards that could be obtained starting from these states. Even if these states are not frequently visited under the optimal policy, we can still explore these states using counterfactual methods to analyze the choices of suboptimal actions. However, the analysis of state transitions typically relies on empirical data collected during the training process. This dependency can limit the model's ability to analyze complex environments, as the model may not have explored all possible states and action combinations. In contrast, methods based on Q-values do not depend on specific state transition details but instead directly evaluate the expected rewards for each state and action combination, effectively overcoming the dependency on empirical data and enhancing their applicability in complex environments.}

In RL, the main difference between the Q-value  and the V-value is that the Q-value involves a combination of state and action, allowing direct quantification of the expected return for each action taken in a given state through Q(s, a). This makes the Q-value particularly well-suited for direct use in the decision-making process, as it provides clear guidance on action selection for each state. In contrast, the V-value is only related to the state and does not involve actions, reflecting the maximum expected return achievable from a state when following the optimal action. This characteristic of not being directly linked to specific actions makes the V-value less straightforward in establishing connections between states from a policy perspective, typically requiring the use of the value functions of subsequent states \( V(s') \) or \( V(s^*) \) to reflect the utility of actions.

\subsection{Average Counterfactual Difference}

To extend the analysis on agent behavior and the expected return (utility) impacts of varying policies, we investigate further into the counterfactual differences by introducing the concepts of Average Counterfactual differences (ACD) for $Q(s,a)$ and $V(s)$. These metrics offer a refined approach to evaluate and compare the efficacy of the optimal action against a spectrum of alternative actions.

For $Q(s,a)$, the \textit{Average Action Counterfactual differences}, denoted as \( \Delta \overline{Q}^{\pi}(s, a^*) \), provides a comprehensive measure by averaging the expected return differences between the optimal action \( a^* \) and all other possible actions within the action space. This metric is particularly useful for evaluating the relative advantage of the optimal action by considering its expected return gain over the average expected return of all alternative actions. The formula for \( \Delta \overline{Q}^{\pi} \) is given by:
\begin{equation}  
\begin{split}
 \Delta \overline{Q}^{\pi}(s, a^*) &= \mathbb{E}_{a \sim A} \left[ \Delta Q^{\pi}(s, a^*, a) \right]\\ &=  \frac{1}{|A|} \sum_{a \in A} \Delta Q^{\pi}(s, a^*, a) \\
  &= \frac{1}{|A|} \sum_{a \in A} \Big[ Q^{\pi}(s, a^*) - Q^{\pi}(s, a) \Big]
\end{split}
\end{equation} 

Similarly, for $V(s)$, \textit{the Average State Counterfactual Differences}, \( \Delta \overline{V}^{\pi}(s) \), quantifies the average expected return change when transitioning from the current state \( s \) to an optimal state \( s^* \), compared to transitions to all other potential states \( s' \) according to policy \(\pi\). This metric aids in understanding the broader implications of policy choices on the agent's position and subsequent expected return. The computation of \( \Delta \overline{V}^{\pi} \) is as follows:
\begin{equation}  
\begin{split}
 \Delta \overline{V}^{\pi}(s) &= \mathbb{E}_{s' \sim S'} \left[ \Delta V^{\pi}(s^*, s') \right]\\ &= \frac{1}{|S'|} \sum_{s' \in S'} \Delta V^{\pi}(s^*, s')\\ &= \frac{1}{|S'|} \sum_{s' \in S'} \Big[  V^{\pi}(s^{*}) - V^{\pi}(s') \Big]
\end{split}
\end{equation}

\noindent \textcolor{black}{where  ${S}' = \{ {s}' \mid   P({s}'|{s},a) \geq 0, \forall a \in A \}$ which means that \( S' \) includes all states that may be reached by taking any action \( a \) in the action space \( A \) from the state \( s \). }

The comparison of \( \Delta \overline{Q} \) with \( \Delta Q \), and \( \Delta \overline{V} \) with \( \Delta V \), highlights their roles in providing analytical depth and breadth within reinforcement learning, crucial for both strategic planning and precise decision-making. More importantly, these models enhance the interpretability and transparency of decision-making processes. ACD offer a broad perspective on the impacts of different actions, aiding in policy clarity, while CD provide detailed insights into specific actions, enhancing operational transparency. Together, they ensure decisions are informed, accountable, and transparent, catering to both overarching polices and immediate actions.

\subsection{Counterfactual Characteristic Value Function}
In the Shapley Value calculations for reinforcement learning, selecting an appropriate characteristic value function is essential. Typically, these functions are derived from fundamental elements like the value function or the action-value function, which can be straightforwardly used to compute what are known as `vanilla' characteristic values. Once the policy is learned, the agent's behavior becomes fixed. In any given state, it almost always chooses the best action, hardly considering other suboptimal options. In this way, choosing the optimal action versus not choosing it forms a counterfactual scenario, which represents theoretical possibilities that do not actually occur. However, to facilitate more detailed comparisons in counterfactuals, particularly between an optimal action and a suboptimal one within the same state, employing counterfactual characteristic value becomes essential. This method thoroughly quantifies the expected return difference when choosing the optimal action instead of a suboptimal alternative, this counterfactual analysis significantly contributes to the transparency and explainability of decisions within reinforcement learning models, as it offers clear insights into the reasons behind preferring certain actions over others, thus making the decision-making process more understandable and justifiable. The CD characteristic value functions are expressed as follows:

\textbf{CD Characteristic Value Function on $\Delta Q$:}
\begin{equation}
v^{Q}_{a^*,a,s}(C) := \Delta Q^C(s^C, a^*, a) 
\end{equation}

\textbf{CD Characteristic Value Function on $\Delta V$:}
\begin{equation}
v^{V}_{{s}^*, {s}'}(C) :=  \Delta V^C({s^C}^*, {s^C}')
\end{equation}

\noindent where \( {s^C}^* \sim P(\cdot|{s}^C,a) \) and \( {s^C}' \sim P(\cdot|{s}^C,a) \) are the states reached from state \( s^C \) by executing the optimal and suboptimal actions, respectively, and \( \Delta Q^C(s^C, a^*, a) \) and \( \Delta V^C({s^C}^*, {s^C}') \) denote the Q-values and V-values obtained by training within a specific subset `C' of the state space.  These values, being derived from a limited set of dimensions, carry different interpretations and consequences than those calculated from the full state space.

The computation of the Shapley Value by CD approach is indispensable for highlighting the relative advantage and direct impact of the optimal action against a specific alternative, offering granular insights into the decision-making process. This method calculates the difference between two specific actions. Using the CD characteristic value function, we can obtain a value called the CD Shapley value, which directly reflects how each feature influences the difference in outcomes between these two actions.

Expanding the scope, the computation of the Shapley Value by ACD approach averages the expected return differences between the optimal action and all suboptimal actions \( A' \), providing a comprehensive evaluation of the action's efficacy across the entire action space. The ACD characteristic value functions are defined as:

\textbf{ACD Characteristic Value Function on $\Delta \overline{Q}$:}
\begin{equation} 
v^{Q}_{a^*,a,s}(C) := \frac{1}{|A|} \sum_{a \in A} \Delta Q^C(s^C, a^*, a) 
\end{equation}

\textbf{ACD Characteristic Value Function on $\Delta \overline{V}$:}
\begin{equation} 
v^{V}_{{s}^*, {s}'}(C) := \frac{1}{|{S^C}'|} \sum_{{s^c}' \in {S^C}'} \Delta V^C({s^C}^*, {s^C}') 
\end{equation}
\textcolor{black}{where ${S^C}' = \{ {s^C}' \mid   P({s^C}'|{s}^C,a) \geq 0, \forall a \in A \}$ which means that \( {S^C}' \) includes all states that may be reached by taking any action \( a \) in the action space \( A \) from the state \( s^C \) of limited dimensions space \(C\).}

The ACD computation approach excels by demonstrating the overall dominance of the optimal action and providing a comprehensive view of the policy's depth and robustness. Unlike the CD method, the ACD method calculates the average difference between all possible actions and a specific action (usually the optimal action). The ACD Shapley value, obtained through the ACD characteristic value function, offers a evaluation on how a specific action generally compares to other available actions.

Together, the CD and ACD computation approach enrich the analysis of policy performance in reinforcement learning by offering both detailed and broad perspectives. While the CD method allows for focused evaluations of specific action dynamics, the ACD delivers a macroscopic understanding of the policy's general utility and resilience. This dual approach strengthens the analytical framework for policy interpretation, emphasizing policy insights over direct policy optimization.

\begin{figure}[!htbp]
\centering
\includegraphics[width=0.97\textwidth]{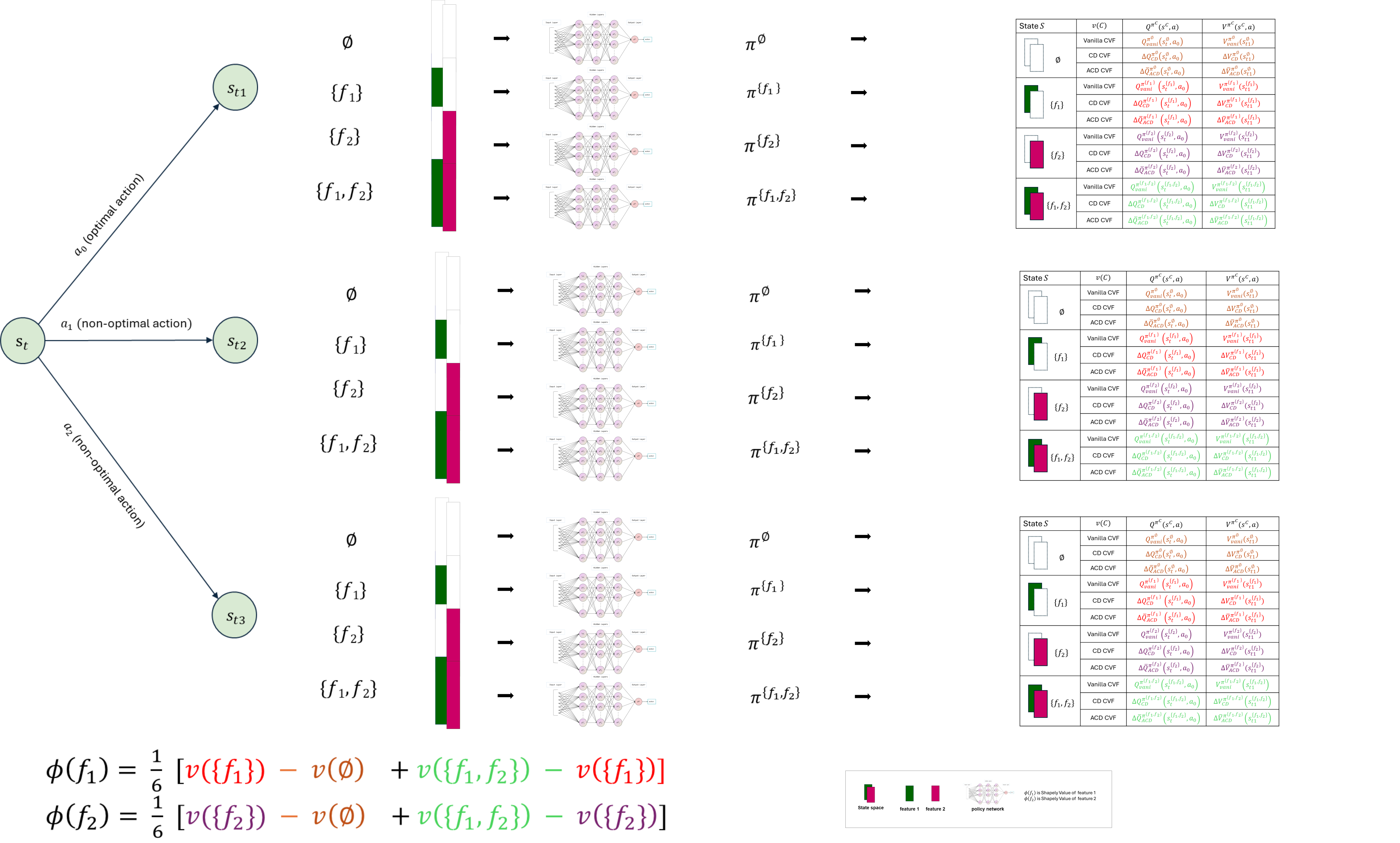}
\caption{Computation of Counterfactual Shapley Value using VANI-CVF, CD-CVF, and ACD-CVF}
\label{fig:Computation_CSV}
\end{figure}
An example of calculating CSV using vani-CVF, CD-CVF, and ACD-CVF, assuming a state with two features, is shown in Fig (\ref{fig:Computation_CSV}). In each CVF, each policy network is independently trained, and the resulting policy $\pi^C$ is learned based on the optimal policy for the current input dimensions $C$.

\section{EXPERIMENTS}
We present experimental results from various domains in RL, applying Counterfactual Shapley Value analysis to understand the influence of optimal versus suboptimal actions on individual dimensions of the state space.
\subsection{Explanation in Gridworld 1 and Gridworld 2}

\begin{figure}[tbhp]
  \centering
  \begin{subfigure}[b]{0.47\linewidth}
    \includegraphics[width=\linewidth]{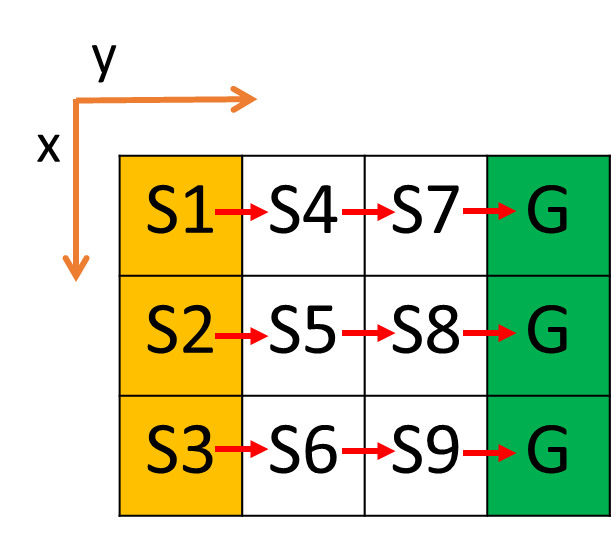}
    \caption{GridWorld-1}
    \label{fig:Gridworld 1}
  \end{subfigure}
  \quad
  \begin{subfigure}[b]{0.47\linewidth}
    \includegraphics[width=\linewidth]{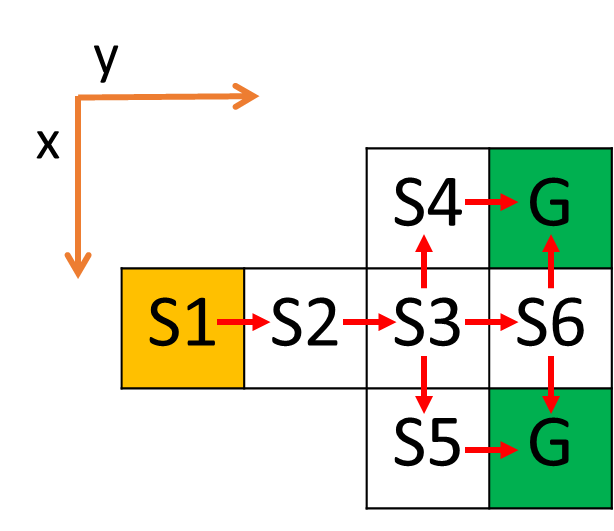}
    \caption{GridWorld-2}
    \label{fig:Gridworld 2}
  \end{subfigure}
  \par\vspace{1ex} 
  
  \begin{subfigure}[b]{0.47\linewidth}
    \includegraphics[width=\linewidth]{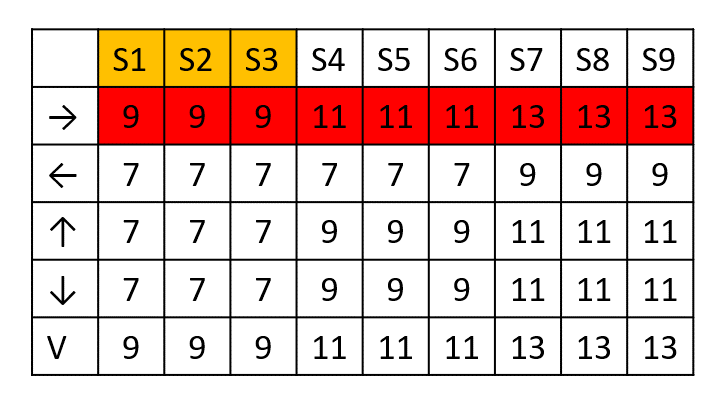}
    \caption{$V(s)$, $Q(s,a)$, and $U(s)$ based on $\pi^*$ in GridWorld-1}
    \label{fig:table 1}
  \end{subfigure}
  \quad
  \begin{subfigure}[b]{0.47\linewidth}
    \includegraphics[width=\linewidth]{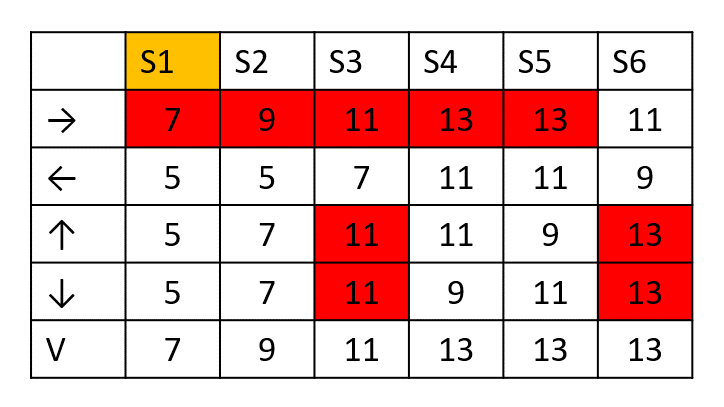}
    \caption{$V(s)$, $Q(s,a)$, and $U(s)$ based on $\pi^*$ in GridWorld-2}
    \label{fig:table 2}
  \end{subfigure}
  
  \caption{Comparison of GridWorld-1 and GridWorld-2, including their corresponding values of $V(s)$, $Q(s,a)$, and $U(s)$ based on $\pi^*$.}
  \label{fig:Gridworld_comparison}
\end{figure}

In \textbf{GridWord-1} on Fig (\ref{fig:Gridworld 1}), it is easily discovered that the optimal action for each non-target state is to move right $\rightarrow$. However, relying solely on the value-table on Fig (\ref{fig:table 1}) generated by state-value functions or action-value functions can only explain the choice of the current action, without explaining the specific impact of that action on the state features. In other words, due to the state possessing multiple dimensions, it is not feasible to directly compare the influence of any dimension on the selection of optimal actions through the analysis of value functions and related techniques.

Using the Counterfactual Difference and Average Counterfactual Difference Shapley Value methods (CD-SPV and ACD-SPV) for calculation of Shapley Value by state-value function (V-value), we can examine the differences between the optimal action and all non-optimal actions, as well as between the optimal action and specific non-optimal actions. This facilitates our comprehension of how these differences affect each dimension of the state. Throughout this process, we will utilize the Shapley Value method offered by the Vanilla characteristic value function (Vanilla-SPV) for thorough analysis.

The findings obtained from the analysis of the \textbf{GridWord-1} scenario using the \textit{ACD-SPV},\textit{CD-SPV} and \textit{Vanilla-SPV} methods are presented in Table (\ref{table:wg1}). The functionalities of these three methods are introduced as follows:

\begin{table*}[htbp]
\centering
\caption{Counterfactual Sharpe Values for Each Dimension of States on GridWord-1}
\small
\begin{tabular}{|c|c|c|c|c|c|c|c|c|c|}
\hline
& $s_1$~:~(0, 0) & $s_4$~:~(0, 1) & $s_7$~:~(0, 2) & $s_2$~:~(1, 0) & $s_5$~:~(1, 1) & $s_8$~:~(1, 2) & $s_3$~:~(2, 0) & $s_6$~:~(2, 1) & $s_9$~:~(2, 2)  \\
\hline
Vani{(0)} & [$3.2$, $7.8$] & [$3.2$, $9.8$] & [$3.2$, $11.8$] & [$4.3$, $6.8$] & [$4.3$, $8.8$] & [$4.3$, $10.8$] & [$4.3$, $6.8$] & [$5.5$, $7.5$] & [$5.5$, $9.5$] \\
\hline
Vani{(1)} & [$3.2$, $5.8$] & [$3.2$, $7.8$] & [$3.2$, $7.8$] & [$4.3$, $4.8$] & [$3.2$, $7.8$] & [$3.2$, $9.8$] & [$5.5$, $3.5$] & [$5.5$, $5.5$] & [$5.5$, $7.5$] \\
\hline
Vani{(2)} & [$3.2$, $5.8$] & [$4.3$, $6.8$] & [$4.3$, $8.8$] & [$3.2$, $5.8$] & [$5.5$, $5.5$] & [$5.5$, $7.5$] & [$5.5$, $3.5$] & [$4.3$, $6.8$] & [$4.3$, $8.8$] \\
\hline
Vani{(3)} & [$4.3$, $4.8$] & [$3.2$, $5.8$] & [$3.2$, $7.8$] & [$5.5$, $3.5$] & [$4.3$, $4.8$] & [$4.3$, $6.8$] & [$4.3$, $4.8$] & [$5.5$, $3.5$] & [$5.5$, $5.5$] \\
\hline
CD(01) & [$0.0$, $2.0$] & [$0.0$, $2.0$] & [$0.0$, $4.0$] & [$0.0$, $2.0$] & [$1.1$, $0.9$] & [$1.1$, $0.9$] & [$-1.2$, $3.2$] & [$0.0$, $2.0$] & [$0.0$, $2.0$] \\
\hline
CD(02) & [$0.0$, $2.0$] & [$-1.1$, $3.1$] & [$-1.1$, $3.1$] & [$1.1$, $0.9$] & [$-1.2$, $3.2$] & [$-1.2$, $3.2$] & [$-1.2$, $3.2$] & [$1.2$, $0.8$] & [$1.2$, $0.8$] \\
\hline
CD(03) & [$-1.1$, $3.1$] & [$0.0$, $4.0$] & [$0.0$, $4.0$] & [$-1.2$, $3.2$] & [$0.0$, $4.0$] & [$0.0$, $4.0$] & [$0.0$, $2.0$] & [$0.0$, $4.0$] & [$0.0$, $4.0$] \\
\hline
ACD & [$-0.4$, $2.4$] & [$-0.4$, $3.0$] & [$-0.4$, $3.7$] & [$-0.1$, $2.1$] & [$-0.1$, $2.7$] & [$-0.1$, $2.7$] & [$-0.8$, $2.8$] & [$0.4$, $2.3$] & [$0.4$, $2.3$] \\
\hline
\end{tabular}
\label{table:wg1}
\end{table*}

\textbf{Vanilla-SPV(i)} or \textbf{Vani(i)}: The list  results show the Shapley Values for each dimension (feature) of the state, which indicate the individual contributions of different dimensions (feature) to the choice of action $i$. In short, these Shapley Values allow us to see the importance and influence of each state dimension in the decision-making process. \textit{Lower index \(i\) corresponds to better actions, with action \(0\) corresponding to the optimal action, which has a Shapley Value for each feature denoted by \( \text{Vani}(0) \). The next best action corresponds to \( \text{Vani}(1) \), and subsequent actions are similarly denoted by \( \text{Vani}(2) \), \( \text{Vani}(3) \), etc.}

\begin{table*}[htbp]
\centering
\caption{Counterfactual Sharpe Values for Each Dimension of State on GridWord-2}
\small
\begin{tabular}{|c|c|c|c|c|c|c|}
\hline
& $s_1$~:~(1, 0) & $s_2$~:~(1, 1) & $s_3$~:~(1, 2) & $s_{4}$~:~(0, 2) & $s_5$~:~(2, 2) & $s_6$~:~(1, 3) \\
\hline
Vanilla{(0)} & [$5.88$, $6.75$] & [$5.88$, $8.75$] & [$4.25$, $10.75$] & [$7.1$, $11.12$] & [$7.1$, $11.15$] & [$7.1$, $11.12$] \\
\hline
Vanilla{(1)} & [$5.88$, $4.75$] & [$5.88$, $6.75$] & [$6.5$, $9.82$] & [$6.5$, $9.82$] & [$6.5$, $9.53$] & [$7.1$, $11.15$] \\
\hline
Vanilla{(2)} & [$5.88$, $4.75$] & [$5.88$, $6.75$] & [$6.5$, $9.53$] & [$6.5$, $9.82$] & [$6.5$, $9.53$] & [$4.25$, $10.75$] \\
\hline
Vanilla{(3)} & [$5.88$, $4.75$] & [$5.88$, $4.75$] & [$5.88$, $6.75$] & [$5.88$, $8.75$] & [$5.88$, $8.75$] & [$5.88$, $8.75$] \\
\hline
CD(0,1) & [$0.0$, $2.0$] & [$0.0$, $2.0$] & [$-2.25$, $0.93$] & [$0.6$, $1.3$] & [$0.6$, $1.62$] & [$0.0$, $-0.03$] \\
\hline
CD(0,2) & [$0.0$, $2.0$] & [$0.0$, $2.0$] & [$-2.25$, $1.22$] & [$0.6$, $1.3$] & [$0.6$, $1.62$] & [$2.85$, $0.37$] \\
\hline
CD(0,3) & [$0.0$, $2.0$] & [$0.0$, $4.0$] & [$-1.63$, $4.0$] & [$1.22$, $2.37$] & [$1.22$, $2.4$] & [$1.22$, $2.37$] \\
\hline
ACD & [$0.0$, $2.0$] & [$0.0$, $2.67$] & [$-2.04$, $2.05$] & [$0.81$, $1.66$] & [$0.81$, $1.88$] & [$1.36$, $0.9$] \\
\hline
\end{tabular}
\label{table:wg2}
\end{table*}

\textbf{CD-SPV{(i,~j)}} or \textbf{CD(ij)}: By examining the differences in the contribution of each dimension of a certain state to the selection of different actions \( i \) and \( j \), the list results show the relative importance of state dimensions (feature) by differences in Shapley Values under varying actions. Since we typically focus on the optimal action \( i =0 \) compared to non-optimal action \( j \), it can describe how the different dimensions of a state contribute differently to the optimal and non-optimal actions. This clarifies why, in certain scenarios, opting for action $i$ over action $j $ might be more advantageous. \textit{CD(0,1) compares the optimal action 0 with the suboptimal action 1, revealing the relative differences in each dimension when action 0 is executed instead of action 1. Conversely, CD(1,0) indicates the comparison when action 1 is executed instead of action 0, and CD(0,1) is the negative of CD(1,0) in each dimension (feature), meaning CD(0,1) = -CD(1,0). The results show that a larger positive difference indicates that the dimension is more important for the decision-making; conversely, a larger negative difference indicates that the dimension has a negative impact on the decision; if the difference is close to zero, it means that this dimension contributes equally under both scenarios. Similarly, CD(0,2) and CD(0,3) compare the optimal action with the second and third suboptimal actions, respectively.}
\textcolor{black}{
We consider all suboptimal actions to be counterfactual actions because, once the policy is learned, the current policy will not execute these suboptimal actions. Of course, in the same state, there may be multiple optimal actions that are equally valid and executable. We choose one of these as the optimal action, and the rest are regarded as counterfactual actions, that is, suboptimal actions.}

\textbf{ACD-SPV} or \textbf{ACD}: By examining the average differences in the contribution of each dimension of a certain state to the selection of the optimal action versus \textbf{all} non-optimal actions, the list results provide the benchmark of contribution for comparing the average difference between executing the optimal and all non-optimal actions. This clarifies the average differential impact across various dimensions on the decision outcomes when optimal actions are executed as opposed to suboptimal actions, analyzing both the quality (positive or negative) and the degree of the impacts.

The Shapley values, computed as the difference between the optimal action and the average of all non-optimal actions, reflect the incremental value of the optimal action. Although these values may appear modest in comparison to the direct contributions of specific actions, the positive values still indicate the superiority of the optimal action over any other, given its positive contribution across all state dimensions. This explains why, in certain situations, selecting the optimal action as opposed to any other action is preferable.

In our analysis of the \textbf{GridWorld-1} environment, several interesting insights were made. When three states \(s_1\), \(s_2\), and \(s_3\) have the same  \(V\) and \(Q\) values after policy learning, we intuitively expect their contributions to deciding the optimal action in each dimension to be consistent. However, when we use the \textit{Vanilla-SPV} method to evaluate the specific contributions of each dimension on optimal action, we find that the contributions vary among similar states. For example, the contributions of state \(s_1\)in two dimensions (features) are 3.2 and 5.8, state \(s_2\) contributes 4.3 and 4.8, while state \(s_3\) contributes 5.5 and 3.5. These results suggest that even though the optimal actions for these states all involve moving to the right (which primarily affects the y-axis coordinate), the \textit{Vanilla-SPV} method shows that in some cases the contribution of the first dimension (x-axis) appears to be higher than the second dimension (y-axis), which opposes our intuitive expectations. This indicates that although the \textit{Vanilla-SPV} method can provide specific contribution data for each dimension, it may not accurately reflect which dimension plays a more critical role in decision-making.

However, when we use the CD-SPV and ACD-SPV methods to evaluate the differences between states when performing optimal and suboptimal actions, we gain deeper insights. Taking state \(s_1\) as an example, its optimal action is to move right (→) while its suboptimal action is to move up (↑) (the other three actions have the same Q or V value, thus they are also considered suboptimal). In this scenario, the differences in dimensions for these two actions are [0, 2]. Similarly, for states \(s_2\) and \(s_3\), the differences in the same conditions are [0, 2] and [-1.2, 3.2] respectively. The reason these two values differ is that the state transition paths are different after executing optimal and suboptimal actions. For example, when executing the optimal action (→), \( s_2 \) transitions to \( s_5 \) and \( s_3 \) transitions to \( s_6 \); whereas when executing the suboptimal action (↑), \( s_2 \) transitions to \( s_1 \) and \( s_3 \) transitions to \( s_2 \). The Shapley Value of each state's features contributing to the decision is different, which results in different Shapley Values for the difference between optimal and suboptimal actions. These results first reveal the differences in each dimension between optimal and suboptimal scenarios, indicating how much each dimension is improved or reduced, and also show that the second dimension contributes more significantly to decision-making. Despite the three suboptimal actions of state \(s_1\) having the same value, their differences vary because performing the optimal action leads \(s_1\) to state \(s_4\), while performing the actions up (↑) and left (→)  keeps it in state \(s_1\), but performing the action down (↓)  moves it to state \(s_2\), demonstrating that even if some actions have the same Q or V value, their utilities can be completely different. ACD-SPV provides a balanced view, showing the average expected return difference between the optimal action and all suboptimal actions. If the counterfactual difference between optimal and suboptimal actions is significantly greater than the average, it indicates a large expected return difference generated by these actions. This can be a useful metric for evaluating similar states in abstract state tasks.

Therefore, from Table (\ref{table:wg1}), it's clear that the contribution of the second dimension significantly exceeds that of the first. The difference in contribution for the second dimension typically ranges between 2 and 3, while the first dimension has little to no positive contribution except in states $s_6$ and $s_9$. This situation occurs because, in these states, the Shapley Value of the first dimension exceeds that of the second, resulting in a slight positive contribution for the first dimension in these specific states. However, this does not alter the fact that the overall contribution of the second dimension is greater than that of the first, clearly indicating that the second dimension contributes more to optimal decision-making than the first.

In \textbf{GridWord-2} on Fig (\ref{fig:Gridworld 2}),
this environment has a starting position and multiple goal states. In states $s_1$ and $s_2$, by applying ACD-SPV and CD-SPV methods, it is clear that the impact of the second dimension on decision-making is significantly greater than that of the first dimension, which can almost be ignored. In states $s_3$ and $s_6$, there are multiple optimal actions. States $s_4$ and $s_5$ are completely symmetrical, and they are equidistant from two different goal states. From the ACD-SPV analysis, we understand that the optimal actions contribute in both the $x$ and $y$ dimensions, but the contribution of the $y$ dimension is more significant, and both contributions are positive. In state $s_6$, the optimal actions primarily involve the first dimension because the optimal actions are to move up (↑) or down (↓), primarily changing the value of the first dimension, moving from (1,3) to either (0,3) or (2,3). In state $s_3$, compared to other actions, the optimal action's contribution in the first dimension is negative and similar in size to the second dimension. This indicates that in state $s_3$, when performing the optimal action, the impact of the first dimension is much less than other options. This is mainly because the forward action changes the $y$ value, having almost no impact on the x-axis, while the right (→) and left (←) actions have a large difference in the contribution of the first dimension, with the former being $4.25$ and the latter being $5.88$, also explaining why in choosing the optimal action, the contribution of the first dimension is much lower than that of other actions.

\subsection{Explanation in FrozenLake}
\textbf{FrozenLake} is a reinforcement learning environment inspired by the challenge of traversing a frozen lake while avoiding treacherous holes, popularized by the OpenAI Gym framework \cite{Brockman2016}. This environment is depicted as a grid where each cell is marked as either safe ice (Squares with Digits) or a perilous hole (H), with designated starting (S) and goal (G) positions. The agent's objective is to navigate from the start to the goal without falling into any holes, guided solely by its position on the grid. The permissible actions are moving north, south, east, or west. The reward system is straightforward: falling into a hole incurs a -10 penalty, reaching the goal yields a +10 reward, and each move costs -1, incentivizing the agent to find the most efficient route to the goal. An episode begins with the agent at the start point and concludes successfully upon reaching the goal or unsuccessfully if the agent falls into a hole.

\begin{figure}[!htbp]
\centering
\includegraphics[width=0.85\textwidth]{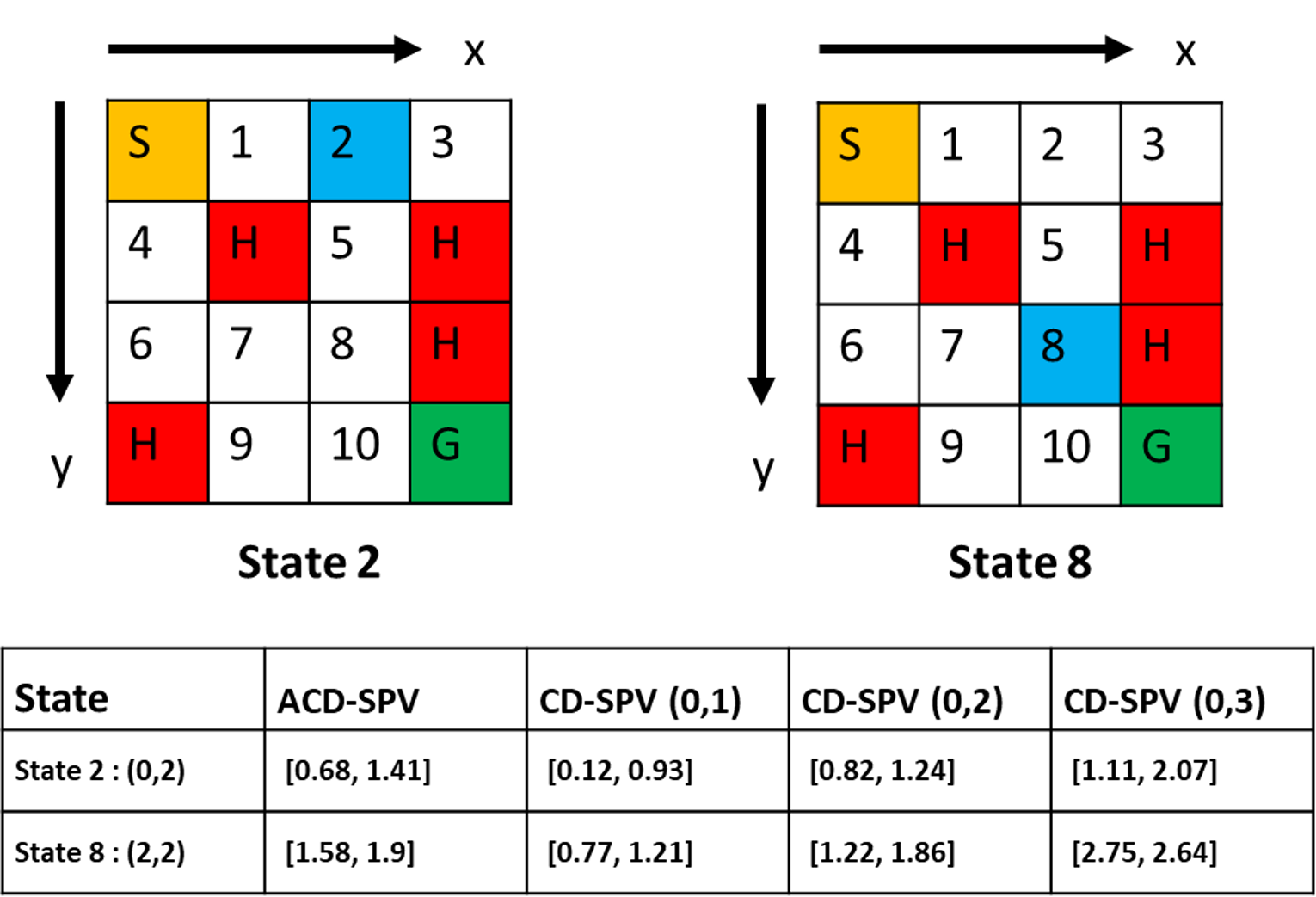}
\caption{State 2 and State 8 are respectively depicted on the upper left and upper right of the figure, representing the coordinates of the agent's current position. State 2 corresponds to [2,0], and State 8 corresponds to [2,2]. Counterfactual Shapley Values for Each Dimension of States on FrozenLake are displayed in the lower half of the figure for these two states.}
\label{fig:cvf_wg_fl}
\end{figure}
\begin{table}[!tbhp]
\centering

\caption{Two states in Taxi for explanation}

\begin{tabular}{|c|c|c|}
\hline
& State 1:[0, 4, 4, 1] & State 2:[0, 0, 0, 1] \\
\hline
ACD & (-1.3, -0.4, -0.02, 0.2) & (-0.9, -0.4, 0.16, -0.0) \\
\hline
CD(0,1) & (-0.1, -0.0, -0.0, 0.2) & (-0.1, -0.0, 0.2, -0.0) \\
\hline
CD(0,2) & (-0.5, -0.1, 0.1, 0.1) & (-0.5, -0.1, 0.0, -0.0) \\
\hline
CD(0,3) & (-2.1, 0.0, 0.0, 0.2) & (-0.0, -0.0, 0.2, -0.1) \\
\hline
CD(0,4) & (-3.9, -1.9, -0.3, 0.2) & (-4.0, -1.9, 0.2, -0.0) \\
\hline
CD(0,5) & (0.1, 0.1, 0.1, 0.2) & (0.1, 0, 0.2, -0.3) \\
\hline
\end{tabular}

\label{table:taxi}
\end{table}

\begin{figure}[!htbp]
\centering
\includegraphics[width=0.85\textwidth]{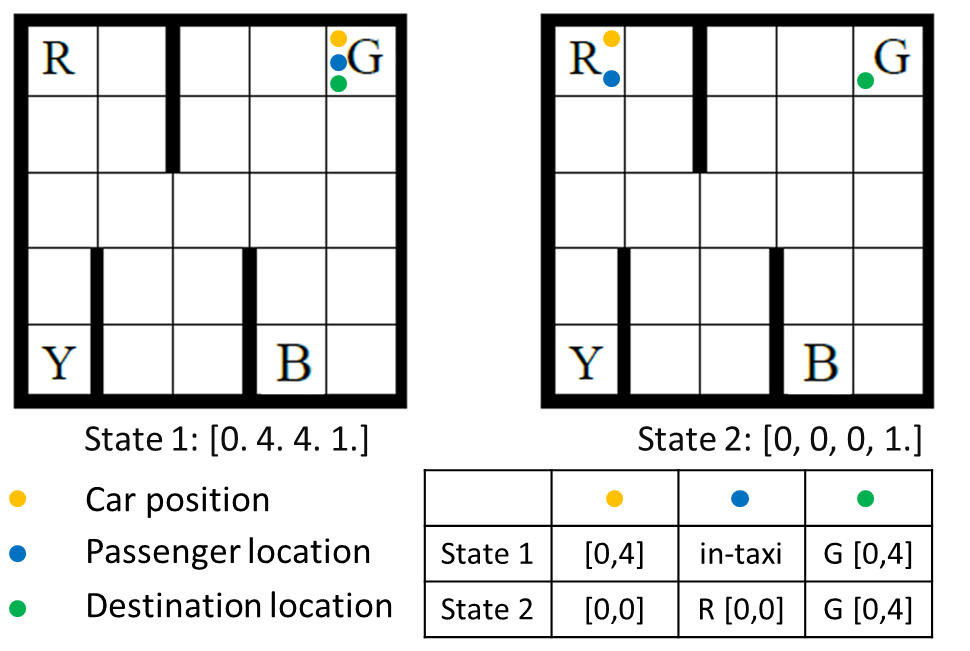}
\caption{State 1 and State 2 represent two different scenarios in the environment. In State 1, the taxi's current position is at [0,4], indicated by the yellow dot, and the passenger is also at position Green, which is [0,4], shown by the blue dot. At this moment, the passenger is inside the taxi, and their destination is also at position Green. Essentially, this state signifies that the passenger has reached the destination but remains in the taxi, and the optimal action for this state is to drop off. Similarly, State 2 is represented as [0,0,0,1], indicating that the taxi is at position [0,0], corresponding to the yellow dot. In this scenario, the passenger, located at position Red, is not inside the taxi, and their destination is at position Green, indicated by the green dot, which is different from the previous positions. In this case, the optimal action for the state is to pick up the passenger.}
\label{fig:Taxi}
\end{figure}

In the \textbf{FrozenLake} environment, when considering the optimal action for State $S_2$ and State $S_8$ in Fig (\ref{fig:cvf_wg_fl}), moving south (↓) is identified as the best action for both two states. This is because moving south (↓) brings one closer to the goal state, proving to be more effective than moving in other directions. For example, in State $S_2$, moving north (↑) would result in staying in the same position ($S_2$), while moving south (↓) would lead to reaching $S_5$. Although the southward move (↓) only has a 0.12 higher contribution in the first dimension compared to moving north (↑), it has a significantly higher contribution of 0.93 in the second dimension, demonstrating its clear advantage. In contrast, moving east (→) or west (←) results in lower contributions in both dimensions, indicating a substantial gap between these decisions and the optimal decision. In the case of State $S_8$, moving south (↓) has a higher contribution towards reaching the goal, especially in the second dimension; moving east (→), however, could result in falling into a hole, thereby creating a significant negative impact in the first dimension. Through such analysis, we can understand why some directions are not the best choices and how these decisions affect the overall policy.

\subsection{Explanation in Taxi}

The \textbf{Taxi} environment, a reinforcement learning benchmark by \cite{Dietterich1998} and featured in OpenAI Gym \cite{Brockman2016}, involves a taxi navigating a 5x5 grid to pick up and drop off a passenger at locations marked Red (R), Green (G), Blue (B), and Yellow (Y). The state captures the taxi's position and the passenger's status, either at a location or in the taxi. The taxi's tasks are to pick up the passenger and deliver them to their destination. It has six actions: move in four directions, pick up, and drop off. Rewards are set to encourage efficiency: -1 for each move, +20 for successful delivery, and -10 for failed pick-up or drop-off, guiding the taxi towards effective passenger transit.

In simple environments, we can easily use the state value function (V-value) to calculate the Shapley Value for interpretation, because in such environments, the transition from one state to another is usually straightforward and simple. This allows us to easily determine the next state and its corresponding value. However, in more complex environments, predicting the next state becomes more challenging, especially if we do not interact with the environment directly. In such cases, we recommend using the action value function (Q-value) to calculate the Shapley Value. Q-values help us assess the expected outcomes of specific actions without actually having to perform those actions. Therefore, Q-values are particularly suitable for use in environments where state transitions are complex or difficult to predict when providing post hoc explanations.

In previous experiments, the primary focus was on actions related to the movement and how various state dimensions (features) influence decision-making, especially regarding movement in different directions. The current research expands the action space to include not only four movement actions but also two critical functions: \textit{picking up} and \textit{dropping off} passengers. This shift allows for an exploration of the decision-making process regarding when to pick up or drop off passengers based on the taxi's current state and its specific location in relation to the passenger and the destination. For example , in State 1, where the taxi and passenger are both at location [0,4], the optimal action is to \textit{drop off} the passenger. Conversely, in State 2, where the taxi and the passenger are at [0,0] but the destination is [0,4], the optimal action is to \textit{pick up} the passenger in Fig (\ref{fig:Taxi}).

In two different states, Table (\ref{table:taxi}) records the counterfactual Shapley Values for the optimal action and other suboptimal actions, reflecting the differences in contributions from different features within the same state to the optimal and suboptimal actions. In State 1, the optimal action is to drop off the passenger, while the worst choice is to pick up the passenger. The other four actions, which are related to movement, rank from second best action to fifth. In State 2, the best action is to pick up the passenger, while the worst choice is to drop off the passenger, with the remaining movement-related actions also ranking from second best action to fifth.

Through the calculation of CD-SPV, we observe that in State 1, the first two dimensions (i.e., the current position of the taxi) contribute less to the decision-making than the movement-related actions. This is because movement itself changes the position of the taxi. This means that in the current state, the dimensions related to location have a smaller impact on the choice of the best action compared to movement actions. From the ACD-SPV results, compared to other actions, only the last dimension (destination location) has a positive contribution, indicating that this dimension is crucial for decision-making in the current state.

Similarly, in State 2, the passenger's location dimension contributes more to the decision to pick up the passenger, the optimal action, than the other dimensions.

\subsection{Explanation in Minesweeper}
Minesweeper includes a 4x4 grid where each cell can display a value of 0, 1, 2, or remain unopened. Each number represents the count of mines adjacent to that cell. Initially, all cells are unopened, and players reveal cells to discover their contents. The game penalizes revealing a mine with a reward of -20, aiming to avoid mines and reveal all non-mine cells. The best possible outcome is a score of 0, achieved by successfully avoiding all mines, with no additional reward for speed or time efficiency. This setup is used for analyzing decision-making in machine learning.

\begin{figure}[!htbp]
\centering
\includegraphics[width=0.85\textwidth]{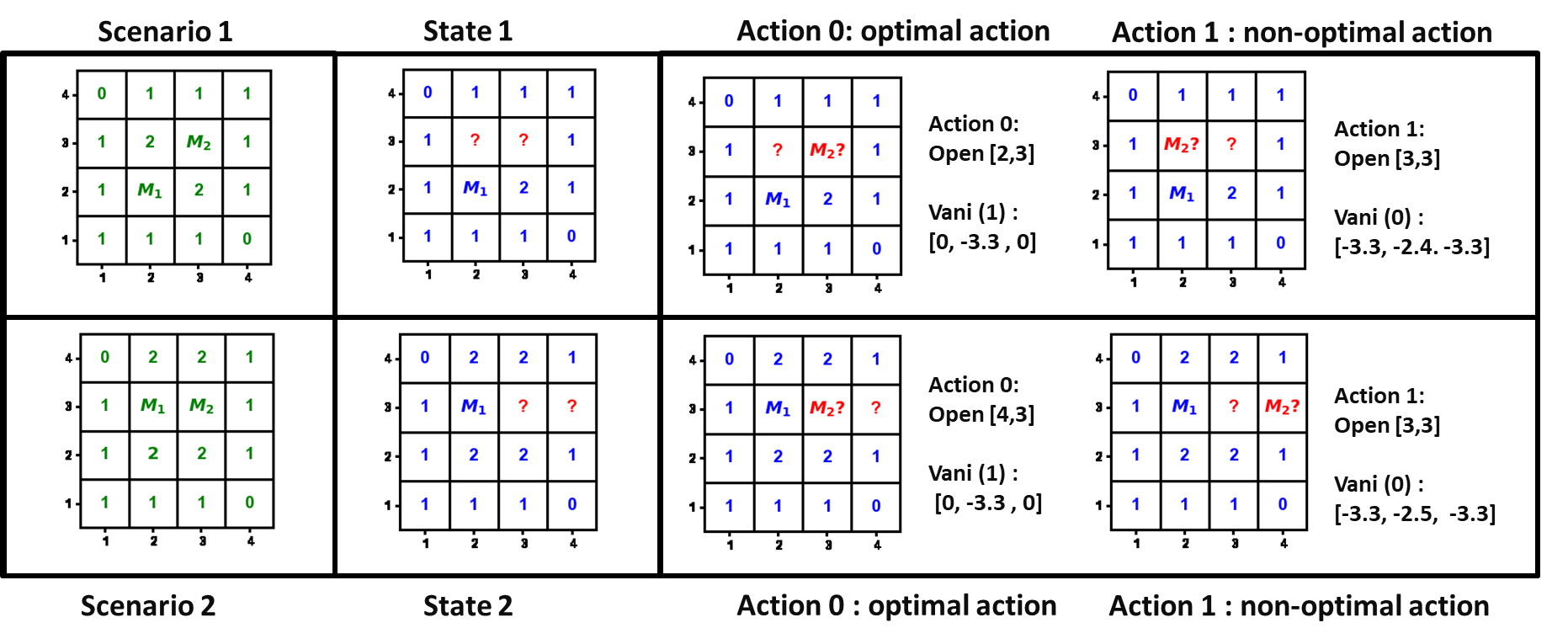}
\caption{The first column shows the actual arrangement of each position in the current game, which is unknown to the player. $M_1$ represents the location of the first mine, $M_2$ represents the location of the second mine, and other numbers indicate the number of mines adjacent to the current position. The second column shows a specific state of the game at the current progress, with blue characters indicating known or observable positions, and red question marks indicating unknown distributions, i.e., positions where mines may be present. The third column lists potential actions, such as $M_2$? indicating the player's assumption of the location of the second mine, i.e., opening another question mark position.}
\label{fig:cvf_min}
\end{figure}

This environment is different from previous ones because here, each selectable action directly corresponds to a specific dimension, and each square's selection is an independent decision in a given state. In Fig \ref{fig:cvf_min}, we demonstrate this with two completely different scenarios. In each scenario, the first cell displays the actual mine distribution, which is usually unknown, but for demonstration purposes, we place it at the forefront. The second cell shows the current state we are studying, and the third cell displays the potential actions, such as predicting the next mine-free location in unopened square. If the prediction is incorrect, resulting in a selection of a position with a mine, the game will end. Otherwise, the game continues or is won.

In State 1, if the mine's location is correctly predicted, the remaining position is safe no matter which one is chosen. However, if the prediction is incorrect, a position with a mine will be selected, leading to game failure. Therefore, the Shapley Values for the two actions are[0,-3,0] , [-3.3,-2.4,-3.3], respectively, Here, the first dimension corresponds to the coordinates of positions [2,3], and the second dimension corresponds to the coordinates of positions [3,3]. All other known dimensions (features) are considered as one dimension (feature), and the overall contribution to the decision-making is combined in the third dimension in Shapley Values. By comparing the Shapley Values of different action, We find that the dimensions without mines and the observable dimensions contribute equally to selecting the optimal action, and if a position with a mine is chosen, both parts will provide a contribution of -3.3 to this decision.  Similarly, in State 2, the optimal action can be determined by comparing the Shapley Values of different actions.

The uniqueness of this environment lies in the fact that each action is directly linked to a specific dimension, allowing us to view them as comparing the effects of different actions. In this comparison, actions with higher Shapley Values indicate that their corresponding dimensions are more important in the current decision-making process. Moreover, the observable dimensions (features) also make significant contributions to the final decision and cannot be overlooked in the decision-making process.

\subsection{Explanation in Pendulum}
The \textbf{Pendulum} environment simulates an inverted pendulum swingup problem 
 from \cite{towers_gymnasium_2023}, where the goal is to apply torque to swing the pendulum into an upright position. The system starts with a pendulum at a random angle ($\theta$) between \([-\pi, \pi]\) and a random angular velocity (torque) between \([-1, 1]\). The observation space consists of the pendulum's x-y coordinates at the free end and its angular velocity,state is $[cos(\theta),sin(\theta),torque]$, while the action space is a single-dimensional array between \([-2, 2]\) representing the applied torque. This setup challenges control algorithms to stabilize the pendulum with minimal effort in a continuous state and action space.
\begin{figure}[!htbp]
\centering
\includegraphics[width=0.42\textwidth]{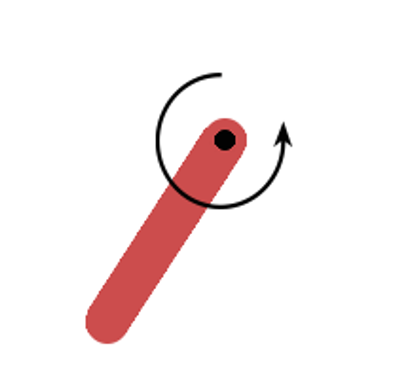}
\caption{State [-0.707, -0.707, 1] of Pendulum}
\label{fig:cvf_Pendulum_env}
\end{figure}

\begin{figure}[!htbp]
\centering
\includegraphics[width=0.85\textwidth]{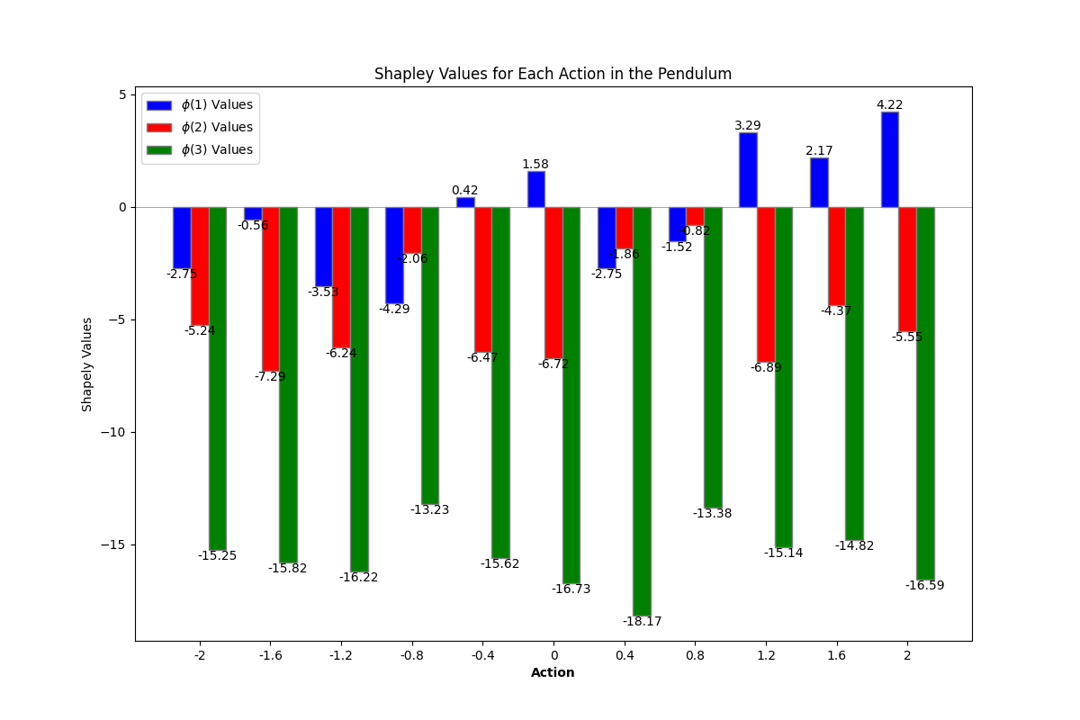}
\caption{Counterfactual Shapley Values for State Dimension Combinations of Pendulum}
\label{fig:cvf_Pendulum}
\end{figure}

Due to the task involving a continuous action space, we first need to convert the continuous action space from [-2, 2] into a discrete action space {-2, -1.6, -1.2, -0.8, -0.4, 0, 0.4, 0.8, 1.2, 1.6, 2} for ease of training and explanation. Next, we choose the state [-0.707, 0.707, 3] for analysis in Fig \ref{fig:cvf_Pendulum_env}. In this state, the pendulum is at a 225-degree angle with an angular velocity of 1, indicating that the pendulum has just fallen from the left side. We calculated the Shapley values for all possible actions in this state and recorded the results in Fig \ref{fig:cvf_Pendulum}. Due to the large action space, we do not show the complete CD-SPV results but directly provide the ACD-SPV results, which are [5.0, -0.76, -1.15]. Analysis shows that the best action is to apply the maximum positive torque. We also find that the first dimension of the state is the most critical for the current decision. Although the third dimension of the state has a significant negative impact (contribution)  on the decision, this impact is common across other action choices and is not distinctive. Therefore, the first dimension of the state is the most important factor in the decision-making process.

\section{Conclusion}

In this research, we introduced a novel application of Counterfactual Shapley Values for enhancing the explainability of reinforcement learning models. Our method not only quantifies the contributions of different dimensions of the state to decision-making but also compares the differences between optimal and non-optimal actions across these dimensions. Simultaneously, it indicates which dimensions contribute positively to optimal actions, which ones negatively, and which ones are crucial features distinguishing optimal from non-optimal actions. Additionally, combining the Shapley Value calculated from the Counterfactual Difference Value Function and the Average Counterfactual Difference Value Function provides more comprehensive details on the contribution of each dimension. The efficacy of the CSV method was demonstrated across multiple RL scenarios, including applications in GridWorld, FrozenLake, and Taxi environments. Each case study provides additional perspectives to aid in understanding the variances in state contributions during the decision-making process and identifying the significant dimensions within the state, offering valuable insights for developers and interested parties seeking to build trust in autonomous systems.

\bibliographystyle{IEEEtran}
\bibliography{IEEEabrv,ref}

\end{document}